\documentclass[letterpaper, 10 pt, journal, twoside]{IEEEtran}  % Comment this line out if you need a4paper

\IEEEoverridecommandlockouts                              % This command is only needed if 
\usepackage{amsmath} % assumes amsmath package installed
\usepackage{amsfonts}
\usepackage{graphicx}
\graphicspath{{figures/}}
\usepackage{subcaption}
\usepackage{array}
\usepackage{multirow}
\usepackage{hyperref}
\usepackage{cite}
\usepackage{comment}
\usepackage[ruled,linesnumbered]{algorithm2e}
\usepackage{tikz}
\usetikzlibrary{positioning}
\usetikzlibrary{arrows}

\newcommand{\reals}{{\mbox{\textbf{R}}}}

\usepackage{xcolor}
\newcommand{\matt}[1]{}
\renewcommand{\matt}[1]{{\color{red} MattJR: {#1}}}

\newcommand{\Ram}[1]{}
\renewcommand{\Ram}[1]{{\color{blue} Ram: {#1}}}
\newcommand{\ram}[1]{}
\renewcommand{\ram}[1]{{\color{blue} Ram: {#1}}}

\newcommand{\myyu}[1]{}
\renewcommand{\myyu}[1]{{\color{cyan} MY: {#1}}}

\newcommand{\XD}[1]{}
\renewcommand{\XD}[1]{{\color{magenta} XD: {(#1)}}}

\newcommand{\yc}[1]{}
\renewcommand{\yc}[1]{{\color{orange} ca: {(#1)}}}

\title{\Large \bf
Stochastic Sampling Simulation for Pedestrian Trajectory Prediction
}

\author{Cyrus~Anderson$^{1}$, Xiaoxiao~Du$^{2}$, Ram~Vasudevan$^{3}$, and Matthew~Johnson-Roberson$^{2}$% <-this % stops a space
\thanks{This work was supported by a grant from Ford Motor Company via the Ford-UM Alliance under award N022884.}
\thanks{$^1$C. Anderson is with the Robotics Institute, University of Michigan, Ann Arbor, MI 48109 USA {\tt\footnotesize andersct@umich.edu}}%
\thanks{$^2$X. Du and M. Johnson-Roberson are with the Department of Naval Architecture and Marine Engineering, University of Michigan, Ann Arbor, MI 48109 USA {\tt\footnotesize xiaodu@umich.edu; mattjr@umich.edu}}%
\thanks{$^3$R. Vasudevan is with the Department of Mechanical Engineering, University of Michigan, Ann Arbor, MI 48109 USA {\tt\footnotesize ramv@umich.edu}}%
}

\begin{document}
\maketitle
% \thispagestyle{empty}
% \pagestyle{empty}

%%%%%%%%%%%%%%%%%%%%%%%%%%%%%%%%%%%%%%%%%%%%%%%%%%%%%%%%%%%%%%%%%%%%%%%%%%%%%%%%
\begin{abstract}
Urban environments pose a significant challenge for autonomous vehicles (AVs) as they must safely navigate while in close proximity to many pedestrians. It is crucial for the AV to correctly understand and predict the future trajectories of pedestrians to avoid collision and plan a safe path. Deep neural networks (DNNs) have shown promising results in accurately predicting pedestrian trajectories, relying on large amounts of annotated real-world data to learn pedestrian behavior.
However, collecting and annotating these large real-world pedestrian datasets is costly in both time and labor. This paper describes a novel method using a stochastic sampling-based simulation to train DNNs for pedestrian trajectory prediction with social interaction. Our novel simulation method can generate vast amounts of automatically-annotated, realistic, and naturalistic synthetic pedestrian trajectories based on small amounts of real annotation. We then use such synthetic trajectories to train an off-the-shelf state-of-the-art deep learning approach Social GAN (Generative Adversarial Network) to perform pedestrian trajectory prediction. Our proposed architecture, trained \textit{only} using synthetic trajectories, achieves better prediction results compared to those trained on human-annotated real-world data using the same network. Our work demonstrates the effectiveness and potential of using simulation as a substitution for human annotation efforts to train high-performing prediction algorithms such as the DNNs.
\end{abstract}

\begin{IEEEkeywords}
Deep Learning in Robotics and Automation, Simulation and Animation
\end{IEEEkeywords}
%%%%%%%%%%%%%%%%%%%%%%%%%%%%%%%%%%%%%%%%%%%%%%%%%%%%%%%%%%%%%%%%%%%%%%%%%%%%%%%%
\section{Introduction}
In crowded urban environments, mobile robots, such as autonomous vehicles (AVs) and social robots, must navigate safely and efficiently while in close proximity to many pedestrians on the road. To avoid collision and ensure a smooth ride, it is crucial for the robots to accurately predict where a nearby pedestrian may move to next. However, the motion and actions of each pedestrian may depend on the  behavior of others, which makes it difficult to forecast \cite{o4luber2010people, alahi2016social, luo2018porca, gupta2018socialgan}. Our goal is to predict all possible future locations and trajectories of pedestrians with probability estimates accounting for social interactions.

Classical approaches to forecasting pedestrian trajectories include Kalman filters, Gaussian Processes~\cite{ellis2009modelling}, and inverse optimal control~\cite{kitani2012activity}, which estimates a model for each pedestrian based on past behavior to forecast the future. These approaches have traditionally focused only on predicting single pedestrians without considering the social interactions between different pedestrians.
These earlier results have been improved upon by extending the frameworks with hand-crafted features to model social interactions~\cite{trautman2010unfreezing}.

More recently, deep neural networks (DNNs) have been used to successfully make long-term pedestrian trajectory prediction accounting for social interactions~\cite{alahi2016social, yi2016pedestrian, gupta2018socialgan, xu2018encoding, nikhil2018convolutional}. Instead of using hand-crafted features, they rely on vast amounts of annotated trajectories to learn social interactions directly from the data. However, it is often difficult, if not impossible, to obtain such large datasets with accurate annotation without resorting to either an enormous effort of manual labeling \cite{pellegrini_eth_data2010improving, drone2016learning} or heavily constructed experiments with instrumented participants~\cite{human32014human3}, both of which are expensive and prone to error. Therefore, it would be valuable to develop a training scheme that requires very small amounts of labeled real-world data, yet still produces satisfactory prediction results for test datasets. 

\begin{figure}[t]
  \centering
  \includegraphics[width=0.5\textwidth]{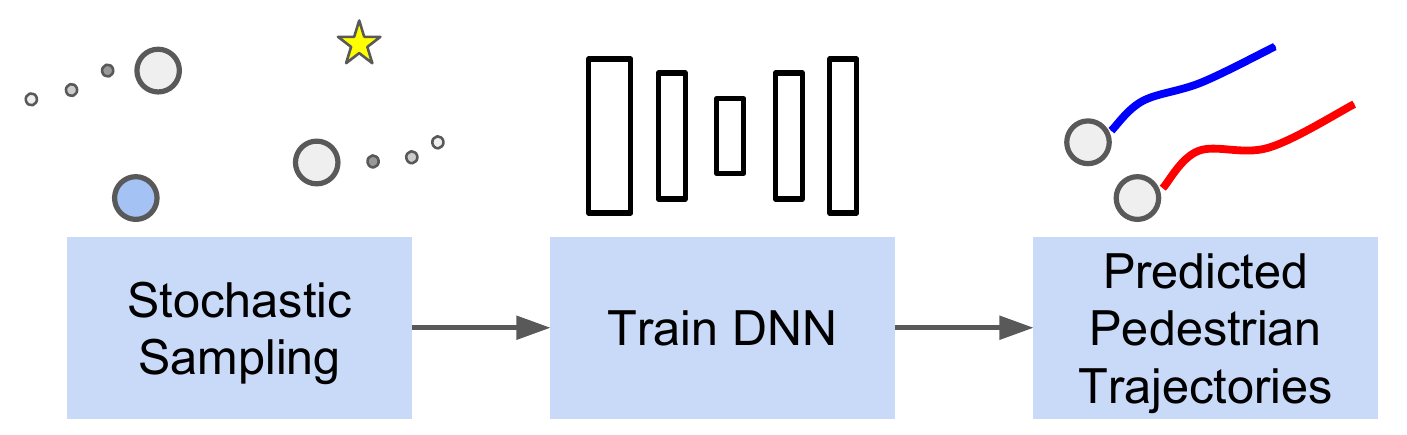}
  \caption{System overview. We propose using a novel stochastic sampling-based simulation system to train a deep neural network (e.g., Social GAN \cite{gupta2018socialgan}) to make socially acceptable pedestrian trajectory predictions.} \label{fig:overview}
\end{figure}

To address this problem, we propose the use of automatically-annotated, realistic pedestrian simulations to train deep neural networks for 2D (top-down view) trajectory prediction. Generating synthetic data for training DNNs has shown promising results for various applications, including image classification~\cite{krizhevsky2012imagenet}, object detection~\cite{johnsonroberson2017matrix}, and pose estimation~\cite{rogez2016mocap}. However, to the best of our knowledge, our work is the first that proposes techniques for synthesizing pedestrian trajectories.

Figure~\ref{fig:overview} shows an overview of our system.
We develop a novel stochastic sampler that can generate tens of thousands of realistic pedestrian trajectories based on limited real-world annotations. We then use these samples to train state-of-the-art DNNs to predict pedestrian trajectories.
In this paper, we use Social GAN~\cite{gupta2018socialgan} as our main prediction network, which provides state-of-the-art predictions and takes into account the interactions of all pedestrians in the scene.
We envision our method being used to train high-performing prediction algorithms (such as DNNs) when there is little annotated data, or where the costs of collecting and labeling real data are high.

Our main contributions include (1) a novel nonparametric model of pedestrians, (2) a method to sample realistic pedestrian trajectories from this model, and (3) experiments training on these sampled trajectories and predicting on benchmark pedestrian datasets, such as the ETH dataset~\cite{o2pellegrini2009you, pellegrini_eth_data2010improving} and the UCY dataset~\cite{lerner2007crowds}.
We demonstrate that Social GAN trained on the realistic sampled pedestrian trajectories alone achieves performance surpassing that achieved by training on real-world human-annotated data.

The paper is organized as follows. Section~\ref{sec:related_works} describes related work in data augmentation and simulation as well as related work in trajectory prediction networks. Section~\ref{sec:methods} describes our proposed simulation model and synthetic trajectory generation process. Section~\ref{sec:experiments} presents pedestrian prediction results on benchmark datasets. Section~\ref{sec:conclusion} concludes this work.

%%%%%%%%%%%%%%%%%%%%%%%%%%%%%%%%%%%%%%%%%%%%%%%%%%%%%%%%%%%%%%%%%%%%%%%%%%%%%%%%
\section{Related Work} \label{sec:related_works}

In this section, we first describe DNN-based pedestrian trajectory prediction methods and our motivation for using the Social GAN. Then, we describe related work in synthetic data generation for training DNNs, such as using physics simulations and domain randomization. Finally, we describe related work in data augmentation and methods for generating synthetic datasets based on real data annotations.

\subsection{DNN-Based Pedestrian Trajectory Prediction Methods}
In recent literature, DNN-based methods, particularly methods based on Long Short-Term Memory (LSTM) networks, have shown successful results in pedestrian trajectory prediction applications \cite{alahi2016social, vemula2018social, fernando2018soft+, gupta2018socialgan, xue2018ss, sadeghian2018sophie,  manh2018scene}. As our method does not consider scene geometry, we refrain from using SS-LSTM \cite{xue2018ss} or scene-LSTM \cite{manh2018scene}, which incorporates scene information. Instead, we select Social GAN \cite{gupta2018socialgan} as our main prediction network, which provides state-of-the-art prediction results and takes into account the interactions of all  pedestrians in the scene. Most recently, a convolutional neural network (CNN)-based trajectory prediction approach was proposed \cite{nikhil2018convolutional}. This approach uses highly parallelizable convolutional layers to handle temporal dependencies and predict future pedestrian positions. However, this CNN-based approach only handles individual trajectory information and does not consider social interactions as Social GAN does. Therefore, in this paper, we chose Social GAN as our basic prediction network.

The Social GAN uses a pooling mechanism together with a Generative Adversarial Network (GAN) to learn the social interactions of pedestrians and produces probabilistic trajectory outcomes. The current Social GAN is a purely data-driven approach \cite{gupta2018socialgan} and benefits from large quantities of annotated trajectory data. However, accurate annotation for large datasets are often difficult, expensive, or impossible to obtain \cite{pellegrini_eth_data2010improving, drone2016learning,human32014human3}. In this paper, we develop a stochastic sampling-based simulation system to automatically generate large amounts of annotated, simulated yet realistic pedestrian trajectories to use as training data for a Social GAN, and aim to produce prediction results comparable or better than the Social GAN prediction results trained on real data.

\subsection{Physics Simulators and Domain Randomization}

Physics simulators and engines can generate new data without the aid of existing data.
Some studies have focused on crafting synthetic data as similar to real data as possible. In~\cite{johnsonroberson2017matrix} and \cite{richter2016playing}, for example, the Grand Theft Auto (GTA) game engine was utilized to produce automatically-annotated, photo-realistic images for object detection and semantic segmentation. In~\cite{kappler2015leveraging}, the  OpenRAVE simulator \cite{diankov2010automated} was used to generate a large-scale database for grasps. This simulated grasps dataset was then used to train a DNN for binary classification (stable or unstable grasps) tasks.

In contrast, domain randomization (DR) methods \cite{tobin2017domain, prakash2018structured} were used to ``bridge the gap'' between simulation and reality. Domain randomization methods focus on bringing variability to the simulation, typically by varying global parameters (e.g., camera pose, shape and number of objects, texture, lighting) and adding noise to the simulated data without much regard to photo-realism. For example, the images synthesized in~\cite{tremblay2018training} contain car models and added geometric objects rendered over random background images. The aim here is to encourage DNNs to learn features invariant to the different kinds of noise added, as well as rendering artifacts and avoid over-fitting. Note that both physics simulations and DR methods do not rely on existing real datasets or annotations.

\subsection{Data Augmentation and (Real-)Data-Driven Synthesis}
Data Augmentation (DA) is another approach to enlarge and enhance training datasets by performing a variety of transformations, typically on images \cite{krizhevsky2012imagenet}. Data augmentation methods usually seek to increase the amount of training data by generating new artificial data from existing real data while preserving label information.
%This process relies on label-preserving transforms, which are generally domain dependent.
Common forms of DA include image translation, horizontal flips/reflections, crops, and perturbations to color (intensity) values \cite{krizhevsky2012imagenet}. 
These methods have been widely used in image classification~\cite{krizhevsky2012imagenet} and object detection~\cite{redmon2016you}.
Other applications include acoustic modeling~\cite{cui2015acoustic} and natural language processing~\cite{xu2016relation_nlp}. So far, we are not aware of any standard label-preserving transforms designed specifically for non-image-based pedestrian trajectories, due in part to the need to account for interactions among pedestrians and scene geometry.

In addition to image transformation, several methods have been proposed to transfer the style between real and synthetic data, adopting the GAN framework \cite{shrivastava2017gaze,sixt2018rendergan,li2018watergan}. Rogez and Schmid~\cite{rogez2016mocap} proposed a synthesis engine to augment existing real images with manual 2D pose labels into 3D poses using 3D Motion Capture (MoCap) data. In a way, these methods were trying to automatically learn the relationship, or transformation, between real and synthetic datasets instead of performing predefined transformation as DA usually does.

Our work is most similar to \cite{rogez2016mocap} in that we also use labeled real-datasets to generate a new, larger synthetic dataset for training.
Unlike~\cite{rogez2016mocap} which synthesizes images for pose estimation, we simulate realistic pedestrian trajectories.
We also perform a variety of perturbations to generate our synthetic data, inspired by DA methods. We show that, with our proposed method, the DNN trained on synthetic data outperforms when trained on real data, even when the synthetic data is generated from a small amount of real annotations.

\section{Stochastic Sampling-Based Simulation} \label{sec:methods}

In this paper, our overall task is to predict future pedestrian trajectories given each pedestrian's previous positions considering social interactions. To do so, we aim to generate large amounts of synthetic pedestrian trajectories for training a Social GAN. A limited amount of real data is given to our simulation system so we can generate realistic trajectories based on how pedestrians actually walk from observed real datasets. In this section, we define the notations and pre-computation steps used in our method and then describe our simulation method in detail.

\subsection{Notations and Pre-computation}
\label{sec:notations}
Let $\mathbf{x}_{tk}$ denote the 2D position (top-down view) at time $t$ for the $k^{th}$ pedestrian, $\mathbf{x}_{tk} \in \reals^2$. Since we utilized a small, real dataset to generate our simulation dataset, we denote $\mathcal{D_R}$ as the given real dataset and $\mathcal{D_S}$ as the generated simulation dataset. In our system, $\left | \mathcal{D_S} \right |  \gg \left | \mathcal{D_R} \right | $, where $\left | \mathcal{D} \right |$ refers to the size of the dataset $\mathcal{D}$. Denote $\mathcal{N}$ as the Gaussian/normal distribution and denote $\mathcal{U}$ as the uniform distribution.

We denote the total number of unique pedestrians in the real dataset $\mathcal{D_R}$ by $K$. At each frame (timestep) in $\mathcal{D_R}$,  we record the number of pedestrians in the scene as $K_p$. $K$ is a known constant for $\mathcal{D_R}$, while $K_p$ may change from frame to frame as pedestrians are entering and exiting the scene. From $K_p$, we can compute the average number of pedestrians in a frame as $\mu_p$ and the variance of the number of pedestrians in the scene as $\sigma_p^2$. 

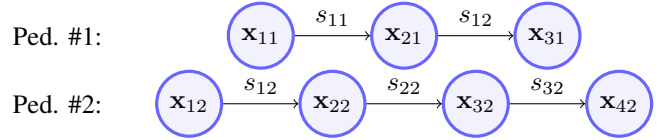
\begin{figure}[h!]
\centering
\begin{minipage}{0.08\textwidth}
  \raggedright
  Ped. \#1:
\end{minipage}
\begin{minipage}{0.4\textwidth}
  \centering
\begin{tikzpicture}[
roundnode/.style={circle, draw=blue!60, fill=blue!5, very thick, minimum size=8mm, text centered},
]
%Nodes
\node[roundnode]      (node3) {$\mathbf{x}_{31}$};
\node[roundnode]      (node2) [left= of node3] {$\mathbf{x}_{21}$};
\node[roundnode]      (node1) [left= of node2] {$\mathbf{x}_{11}$};
 
%Lines
\draw[->] (node1) -- node [above] {$s_{11}$}  (node2);
\draw[->] (node2) -- node [above] {$s_{12}$} (node3);
\end{tikzpicture}
\end{minipage}
\begin{minipage}{0.08\textwidth}
  \raggedright
  Ped. \#2:
\end{minipage}
\begin{minipage}{0.4\textwidth}
  \centering
\begin{tikzpicture}[
roundnode/.style={circle, draw=blue!60, fill=blue!5, very thick, minimum size=8mm, text centered},
]
%Nodes
\node[roundnode]      (node4) {$\mathbf{x}_{42}$};
\node[roundnode]      (node3) [left= of node4] {$\mathbf{x}_{32}$};
\node[roundnode]      (node2) [left= of node3] {$\mathbf{x}_{22}$};
\node[roundnode]      (node1) [left= of node2] {$\mathbf{x}_{12}$};
 
%Lines
\draw[->] (node1) -- node [above] {$s_{12}$}  (node2);
\draw[->] (node2) -- node [above] {$s_{22}$}  (node3);
\draw[->] (node3) -- node [above] {$s_{32}$}  (node4);
\end{tikzpicture}
\end{minipage}
\caption{An illustration for computing pedestrian trajectory speed. Suppose we observed two pedestrians in $\mathcal{D_R}$; pedestrian 1 appeared in a sequence of three timesteps ($T_1=3$), pedestrian 2 appeared in a sequence of four timesteps ($T_2=4$). The $\mathbf{x}_{tk}$ denotes the X-Y position at each timestep $t$ for pedestrian $k$. The speed at each timestep can be calculated using \eqref{eqn:speedeach}.}
\label{fig:illustration_speed}
\end{figure}

From $\mathcal{D_R}$, we can also compute the walking speed for each pedestrian, following   
\begin{align} \label{eqn:speedeach}
   s_{tk} = \frac{||\mathbf{x}_{(t+1)k} - \mathbf{x}_{tk}||}{\Delta t},
\end{align}
where $\mathbf{x}_{tk}$ denotes the 2D position at each timestep $t$ for pedestrian $k= 1,...,K$, $\Delta t$ is the difference in time between two frames/timesteps (fixed), and $||\cdot||$ denotes Euclidean distance. Figure~\ref{fig:illustration_speed} shows a simple illustration for computing the speed for two pedestrians. Note that a pedestrian in $\mathcal{D_R}$ may appear in sequences of varying lengths (due to entering and exiting the scene or due to available data). We denote the sequence length (length of observed timesteps) as $T_k$ for pedestrian $k$ in $\mathcal{D_R}$. Also, note that the speed $s_{tk}$ can vary in each step for real pedestrians.

Given the speed for each pedestrian at each timestep, we can compute the average walking speed for $k^{th}$ pedestrian as $\bar{s}_k$. In our method, we assume that all persons walk with the same speed variation between timesteps and we compute $\sigma_s^2$ as the pooled variance across all $s_{tk}, \forall t, k$. Note that $\bar{s}_k$ changes for each pedestrian and $\sigma_s^2$ is the same for all pedestrians. 

The above summary statistics reflect how pedestrians walk in the real dataset and are used later to generate synthetic trajectories in our sampling scheme.

\subsection{Sampling Number of Pedestrian and Walking Speed}
\label{sec:sampling_speed}
In our simulation, we use stochastic sampling to generate realistic pedestrian trajectories. In this section, we describe the method to sample the number of pedestrians and walking speeds for the simulated dataset.

Let $n_p$ denote the number of pedestrians in a frame in the simulated dataset. In simulating a single set of pedestrians, we sample $n_p$ based on statistics from the given real dataset. We already obtained the average number of pedestrian in a frame $\mu_p$ and the variance of the number of pedestrians in the scene $\sigma_p^2$ from Section~\ref{sec:notations}. We assume the number of pedestrians at each time follows a normal distribution $\mathcal{N}(\mu_p, \sigma_p^2)$  left-truncated at zero, which we denote with $\mathcal{N}(\mu_p, \sigma_p^2, 0)$. We can then sample $n_p \sim \mathcal{N}(\mu_p, \sigma_p^2, 0)$.

Regarding walking speed,  we model each pedestrian as walking at a desired constant speed. Denote $s^{(i)}$ as the speed of the $i^{th}$ sampled pedestrian. We use the superscript to distinguish between the index of the stochastically sampled and real datasets. For each $i$, we first uniformly sample an average speed value, $\bar{s}^{(i)}$, from the pool of average speeds from real pedestrians, denoted as  $\bar{s}^{(i)} \sim \mathcal{U}(\{\bar{s}_k\}_{k=1}^K)$. The variance of speed is assumed to be the same as real pedestrians, $\sigma_s^2$. Then, we can sample $s^{(i)}$ based on a truncated normal distribution, $\mathcal{N}(\bar{s}^{(i)}, \sigma_s^2, 0)$.

\subsection{Pedestrian Trajectory Sampling}
\label{sec:sampling}

Based on the number of pedestrians and walking speeds sampled above, we can determine the actual paths of the pedestrians. We generate the pedestrian trajectories by assigning the sampled speeds to these paths.

We represent the real dataset as a collection of trajectories, i.e., $\mathcal{D_R} = \{f_k\}_{k=1}^K$, where $f_k$ is the trajectory for the $k^{th}$ pedestrian. The trajectory $f_k = \{ \mathbf{x}_{tk} | t = t_k, ...,T_k\}$, for pedestrian $k$ present in the scene from time $t_k$ to $T_k$.

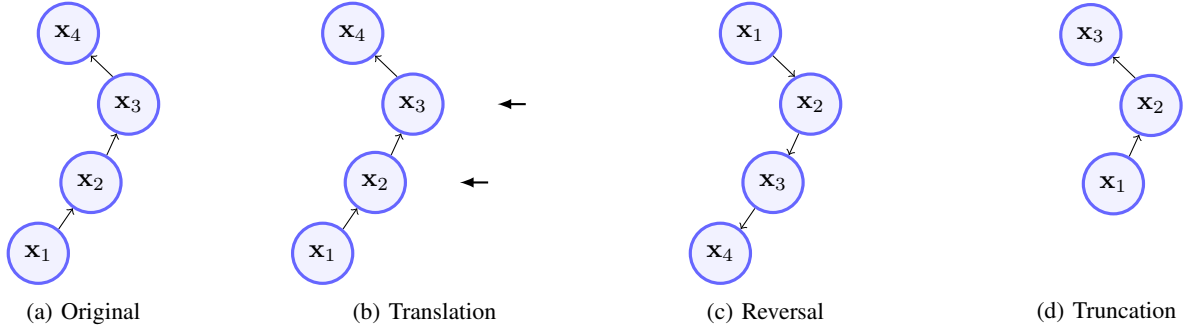
\begin{figure*}[h!]
\centering
\begin{subfigure}{.25\textwidth}
  \centering
\begin{tikzpicture}[
roundnode/.style={circle, draw=blue!60, fill=blue!5, very thick, minimum size=8mm, text centered},
]
%Nodes
\node[roundnode]      (node4)                              {$\mathbf{x}_4$};
\node[roundnode]      (node3) [below=0.1cm of node4, xshift=0.8cm] {$\mathbf{x}_3$};
\node[roundnode]      (node2) [below=0.2cm of node3, xshift=-0.5cm] {$\mathbf{x}_2$};
\node[roundnode]      (node1) [below=0.1cm of node2, xshift=-0.7cm] {$\mathbf{x}_1$};
 
%Lines
\draw[->] (node1.50) -- (node2.235);
\draw[->] (node2.60) -- (node3.248);
\draw[->] (node3.120) -- (node4.315);
\end{tikzpicture}
  \caption{Original}
%   \label{fig:sfig1}
\end{subfigure}%
\begin{subfigure}{.25\textwidth}
  \centering
\begin{tikzpicture}[
roundnode/.style={circle, draw=blue!60, fill=blue!5, very thick, minimum size=8mm, text centered},
ghostnode/.style={circle, draw=blue!0, fill=blue!0, very thin, minimum size=8mm},
]
%Nodes
\node[ghostnode]      (ghost)                              {};
\node[roundnode]      (node4) [left=1.5cm of ghost, yshift=-0.0cm] {$\mathbf{x}_4$};
\node[roundnode]      (node3) [below=0.1cm of node4, xshift=0.8cm] {$\mathbf{x}_3$};
\node[roundnode]      (node2) [below=0.2cm of node3, xshift=-0.5cm] {$\mathbf{x}_2$};
\node[roundnode]      (node1) [below=0.1cm of node2, xshift=-0.7cm] {$\mathbf{x}_1$};
 
%Lines
\draw[->] (node1.50) -- (node2.235);
\draw[->] (node2.60) -- (node3.248);
\draw[->] (node3.120) -- (node4.315);
\draw[arrows={latex-}, thick, shorten <=20] (node3) --++ (1.5,0);
\draw[arrows={latex-}, thick, shorten <=20] (node2) --++ (1.5,0);
\end{tikzpicture}
  \caption{Translation}
%   \label{fig:sfig1}
\end{subfigure}%
\begin{subfigure}{.25\textwidth}
  \centering
\begin{tikzpicture}[
roundnode/.style={circle, draw=blue!60, fill=blue!5, very thick, minimum size=8mm, text centered},
]
%Nodes
\node[roundnode]      (node4)                              {$\mathbf{x}_1$};
\node[roundnode]      (node3) [below=0.1cm of node4, xshift=0.8cm] {$\mathbf{x}_2$};
\node[roundnode]      (node2) [below=0.2cm of node3, xshift=-0.5cm] {$\mathbf{x}_3$};
\node[roundnode]      (node1) [below=0.1cm of node2, xshift=-0.7cm] {$\mathbf{x}_4$};
 
%Lines
\draw[<-] (node1.50) -- (node2.235);
\draw[<-] (node2.60) -- (node3.248);
\draw[<-] (node3.120) -- (node4.315);
\end{tikzpicture}
  \caption{Reversal}
%   \label{fig:sfig1}
\end{subfigure}%
\begin{subfigure}{.25\textwidth}
  \centering
\begin{tikzpicture}[
roundnode/.style={circle, draw=blue!60, fill=blue!5, very thick, minimum size=8mm, text centered},
ghostnode/.style={circle, draw=blue!0, fill=blue!0, very thin, minimum size=8mm},
]
%Nodes
\node[roundnode]      (node4)                              {$\mathbf{x}_3$};
\node[roundnode]      (node3) [below=0.1cm of node4, xshift=0.8cm] {$\mathbf{x}_2$};
\node[roundnode]      (node2) [below=0.2cm of node3, xshift=-0.5cm] {$\mathbf{x}_1$};
\node[ghostnode]      (node1) [below=0.1cm of node2, xshift=-0.7cm] {};
 
%Lines
\draw[->] (node2.60) -- (node3.248);
\draw[->] (node3.120) -- (node4.315);
\end{tikzpicture}
  \caption{Truncation}
%   \label{fig:sfig1}
\end{subfigure}%

\caption{An example of the perturbations used to sample pedestrian paths. Suppose (a) is a sample trajectory from real data, (b) represents the ``translation'' perturbation, with the arrows representing horizontal displacement. (c) is an example of the ``reversal'' perturbation where the start and final locations as well as waypoints are flipped. (d) shows an example of ``truncation'' perturbation where the last location was truncated.} \label{fig:perturbations}
\end{figure*}

For each sampled pedestrian $i$, we first uniformly sample a trajectory, ${f}^{(i)}$, from the pool of all real pedestrian paths, denoted as  $f^{(i)} \sim \mathcal{U}(\{f_k\}_{k=1}^K)$. Then, we apply the following three types of perturbations to the  $f^{(i)}$:
\begin{itemize}
    \item Translation by an amount $\Delta\mathbf{x} \sim \mathcal{U}([-r, r] \times [-r, r])$, where $r$ is the user-defined displacement in each axis of the 2D plane.
    \item Reversal with probability $p_r$: We reverse the start and ending locations as well as all the waypoints in between.
    \item Truncation by a random number of steps.
\end{itemize}
Fig~\ref{fig:perturbations} shows an illustration for the perturbations. The pedestrian then follows the path $g$, a piecewise linear spline fit \cite{loan1999introduction} to the perturbed $f^{(i)}$. All synthetic pedestrians have fixed $N+1$ timesteps, and we denote a sampled set of trajectories as the set  $\mathcal{X^S} = \{\mathbf{x}_{li}| i=1,..., n_p, l=1,...,N+1 \}$, for $N+1$ timesteps and $n_p$ pedestrians in the scene.  

We run Algorithm~\ref{algo:sampler} for a user-defined $M$ number of times to generate the entire large-scale simulated trajectory dataset.
In practice we see that $M > 20$ produces datasets yielding competitive models, with better performance for larger $M$.

\begin{algorithm}[h!]
\SetAlgoLined
\KwIn{$\mathcal{D_R} = \{f_k\}_{k=1}^K$, $\{\bar{s}_k\}_{k=1}^K$, $N$, $\Delta t$, $\mu_p, \sigma_p^2$, $\sigma_s^2$}  
\KwOut{$\mathcal{X^S}$}
$n_p \sim \mathcal{N}(\mu_p, \sigma_p^2, 0)$  \\
\ForEach{$i = 1,...,n_p$}{
  $\bar{s}^{(i)} \sim \mathcal{U}(\{\bar{s}_k\}_{k=1}^K)$  \\
    $s^{(i)} \sim \mathcal{N}(\bar{s}^{(i)}, \sigma_s^2, 0)$   // see Section~\ref{sec:sampling_speed}  \\
  $f^{(i)} \sim \mathcal{U}(\{f_k\}_{k=1}^K)$         \\
  $\tilde{f}^{(i)} \leftarrow \mathrm{peturb}(f^{(i)})$         // see Section~\ref{sec:sampling}  \\
  $g \leftarrow \mathrm{spline}(\tilde{f}^{(i)})$             \\
  \ForEach{$l=1,...,N+1$}{
    $\mathbf{x}_{li} \leftarrow g(s^{(i)}\Delta t l)$
    }
  }
\caption{Stochastic Sampling for Pedestrian Trajectory Simulation}
\label{algo:sampler}
\end{algorithm}

\section{Experiments} \label{sec:experiments}
To measure the effectiveness of the proposed method, we first generated synthetic datasets from the sampling-based simulation method described above. We employ Social GAN~\cite{gupta2018socialgan}, a state-of-the-art deep learning trajectory prediction network architecture, to perform a prediction based on our simulated data. We evaluate our method on two widely used pedestrian trajectory datasets. The ETH dataset~\cite{o2pellegrini2009you, pellegrini_eth_data2010improving} contains over 850 labeled frames of data in each of two distinct scenes, \textit{ETH} and \textit{Hotel}. The UCY dataset~\cite{lerner2007crowds} also has two scenes, \textit{Zara} and \textit{University}, each with over 1500 labeled frames.
Using leave-one-out cross validation, we evaluated the Social GAN's prediction outputs on each scene, having trained on data from the other three scenes.

\subsection{Baselines}
Since the focus of this paper is synthetic trajectory dataset generation, we compare Social GAN prediction results trained on the following four methods:

\begin{itemize}
  \item \textbf{Real:} Train on all the available frames of real data in each scene (approximately 5,000 frames in total).
  \item \textbf{Synth-Large:} Train on a large synthetic dataset. We sampled simulated trajectories 500 times ($M=500$) using Algorithm~\ref{algo:sampler} with $N=20$ timesteps for each scene. When sampling from \textit{University} dataset, we  sampled $M=100$ times due to the large numbers of pedestrians in the scene. This yields over 20,000 simulated, labeled frames in every train-test cross validation split.
  \item \textbf{Synth-Equal:} Train on a synthetic dataset that has the same size as the real dataset (which is much smaller than the size of Synth-Large).  We sampled simulated trajectories such that the number of frames of simulated pedestrian trajectories is equal to that of the real data for each scene.
  \item \textbf{Real + Synth-Large:} Train on the combined data from Real and Synth-Large. This is to evaluate the effect of including real data in training.
\end{itemize}

The Synth-Large dataset has over 15k more frames in in each cross validation split than using 100\% of real data (``Real-100\%''). In addition, we examined the effect of using an even smaller number of labeled frames of real data. We randomly selected 20\% of the real data from each scene (around 1.2k frames) and used this to train the Social GAN. We also generated a separate set of synthetic datasets based only on the 20\% real data and reported prediction results as well. These results were reported under ``20\%'' columns in Table~\ref{table:main_results}.

\subsection{Evaluation Metrics}
Since Social GAN makes probabilistic predictions, we treat the predicted position for the $i^{th}$ pedestrian at timestep $t$ as a random variable $\mathbf{y}_{ti}$. To thoroughly sample the predictive distribution for $\mathbf{y}_{ti}$, we made 100 probabilistic predictions of the pedestrian's possible locations. We denote the ground truth position as $\mathbf{x}_{ti}$.
Similar to prior work \cite{alahi2016social, gupta2018socialgan}, we used the following error metrics:

\subsubsection{Average Displacement Error (ADE)}  Expected distance between the ground truth (GT) pedestrian location and the probabilistic prediction. We estimate this for the dataset by averaging across all $N_p$ pedestrians in the dataset and all predicted timesteps ($t=1,...,T$) as
\begin{equation}
ADE = \frac{1}{N_p \times T}\sum_{i=1}^{N_p}\sum_{t=1}^{T} \mathbb{E}[||\mathbf{y}_{ti}-\mathbf{x}_{ti}||],
\end{equation}
where $\mathbb{E}[\cdot]$ refers to the expected value.

\subsubsection{Minimum Displacement Error (MDE)} Minimum distance between the GT pedestrian location and our predictions, averaged across pedestrians and timesteps. Denoting the $j$th probabilistic prediction by $\mathbf{y}_{ti}^{(j)}$, the MDE is given by
\begin{equation}
MDE = \frac{1}{N_p \times T}\sum_{i=1}^{N_p}\sum_{t=1}^{T} \min_j\{||\mathbf{y}_{ti}^{(j)}-\mathbf{x}_{ti}||\}.
\end{equation}

\subsubsection{Final Displacement Error (FDE)} Expected distance between the GT pedestrian location at the final time step $T$ and the predicted final position. This is averaged across pedestrians, written as
\begin{equation}
FDE = \frac{1}{N_p}\sum_{i=1}^{N_p} \mathbb{E}[||\mathbf{y}_{Ti}-\mathbf{x}_{Ti}||].
\end{equation}

The ADE provides a measure of spread in the predictions, in that a model producing a large spread will necessarily have a large ADE. Unless the model makes predictions near the pedestrian, a small spread in predictions will not ensure a low ADE.
The MDE reflects recall in the ``best case'', measuring the closest prediction to the pedestrian.
The FDE is equivalent to ADE measured only at the final timestep. Ideally, we want the ADE, MDE, and FDE to all have low values to show that all the predictions are close to each pedestrian GT location.

\subsection{Training Parameters}
The Social GAN network architecture is trained with a learning rate of 0.001 and batch sizes of 64 for 200 epochs, following the training procedure in~\cite{gupta2018socialgan} for each experiment. In alignment with this, we use a timestep of $\Delta t = 0.4$ s when sampling pedestrian trajectories. Predictions are made by observing pedestrian trajectories for 8 timesteps (3.2 s) and making predictions for the next 8 timesteps ($T=8$).

For each scene we separately calculate the mean and standard deviation for the number of pedestrians at each timestep ($\mu_p$ and $\sigma_p$), and the standard deviation about their desired speeds ($\sigma_s$). These values are also used when sampling from the smaller amounts of real data, as these can be reliability estimated with much less effort than that needed to build a dataset. The summary statistics calculated for each scene are given in Table~\ref{table:summary_statistics}. Note that the \textit{University} scene alone (``Univ'' row) contains larger mean and variance for number of pedestrians compared with the rest.

\begin{table}[h!]
\caption{Summary statistics for each scene.}
\label{table:summary_statistics}
\centering
\begin{tabular}{ |c|c|c|c| } 
\hline
Dataset & $\mu_p$ & $\sigma_p$ & $\sigma_s$ \\
\hline
ETH & 6.15 & 4.46 & 0.35 \\
Hotel & 5.60 & 3.41 & 0.15 \\
Zara & 7.36 & 3.95 & 0.25 \\
Univ & 26.77 & 20.31 & 0.27 \\
\hline
\end{tabular}
\end{table}

\subsection{Comparison of Prediction Performance}
\begin{figure*}[h]
\includegraphics[width=0.9\textwidth]{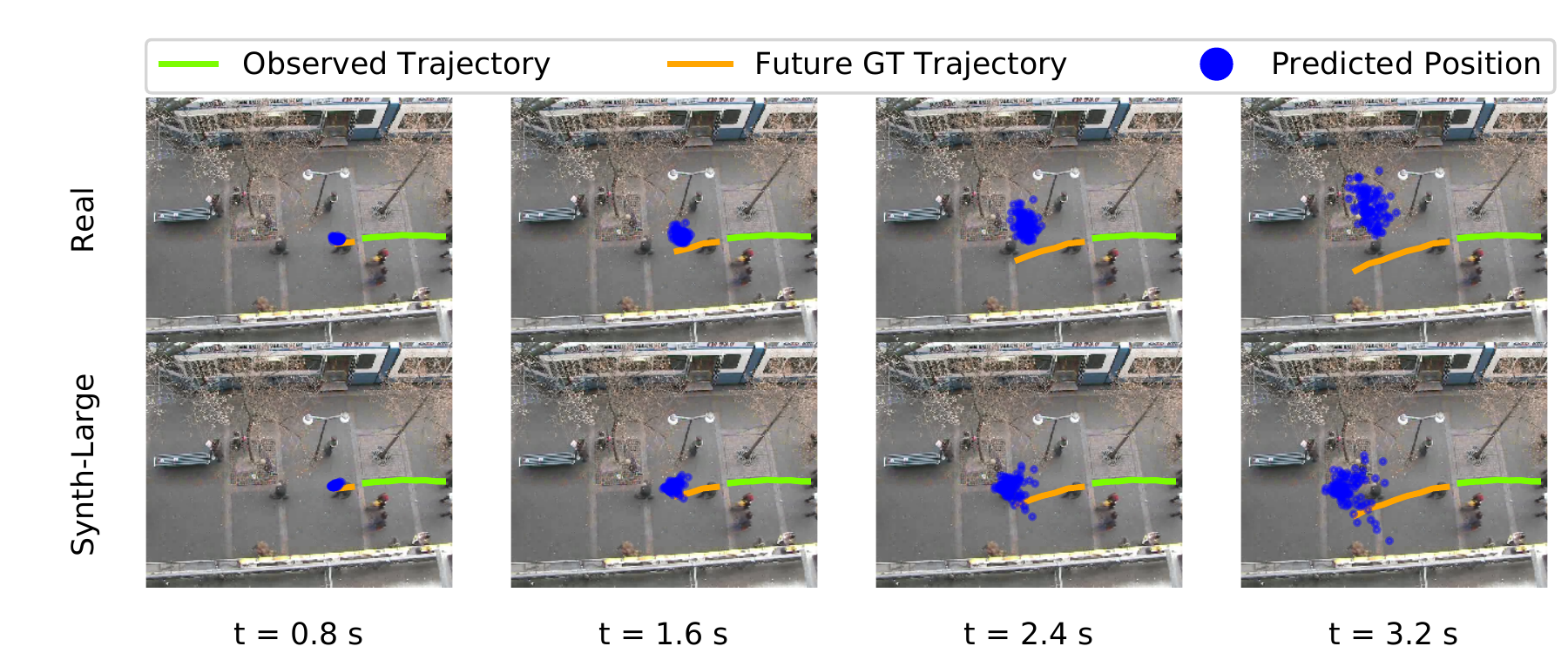}
\includegraphics[width=0.9\textwidth]{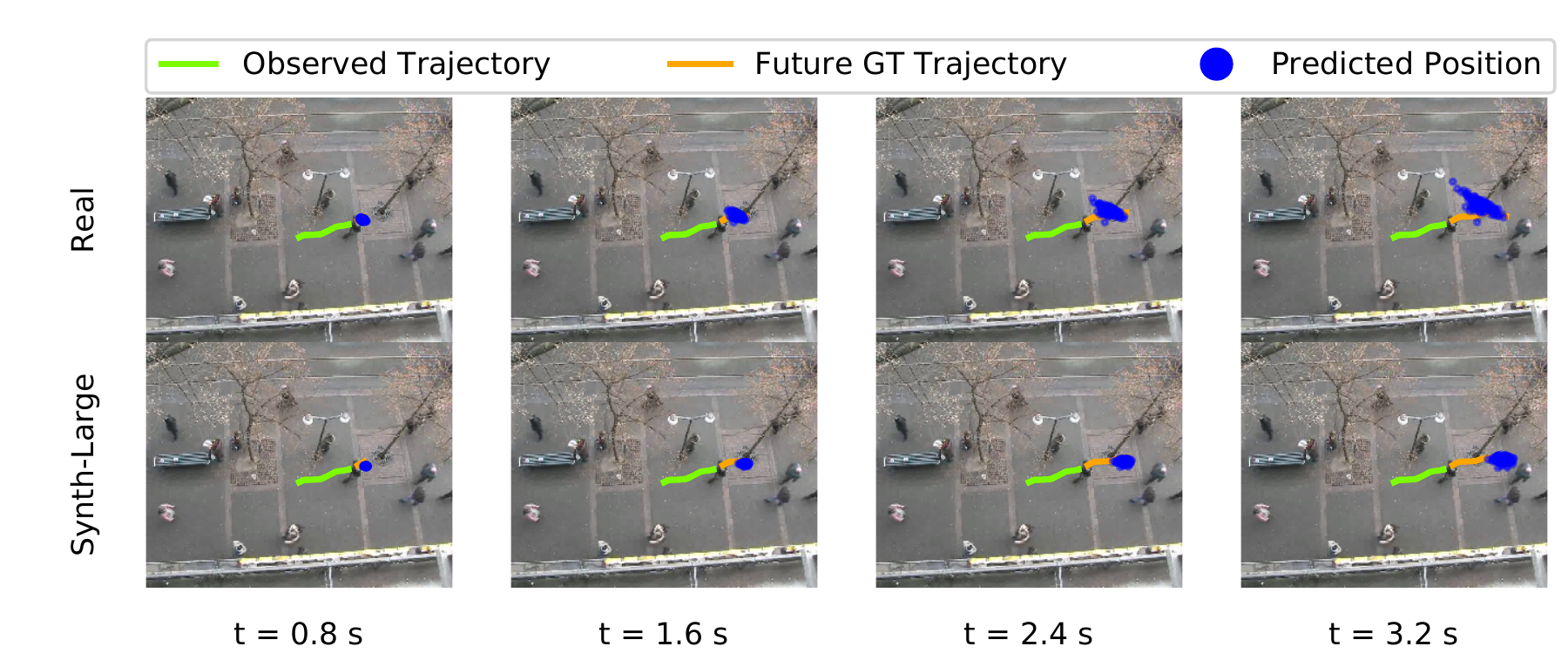}
\centering
\caption{Two examples of prediction results from Social GAN trained on ``Real-100\%'' (100\% real data only) and on ``Synth-Large-100\%'' (large synethic data sampled from all real data).  We show the pedestrian's predicted position for 2, 4, 6, and 8 timesteps into the future. The Social GAN jointly predicts future positions for all timesteps. The green solid line represents the observed trajectory (in the past). The orange solid line represents the ground truth trajectory for future timesteps. The blue dots represent the 100 probabilistic predictions of pedestrian locations. The ''Synth-Large'' predictions are closer to the ground truth position and can obtain decreased error across all evaluation metrics. } \label{fig:predictions}
\end{figure*}

Table~\ref{table:main_results} shows the ADE, MDE, and FDE comparison results across all cross validation datasets, predicted using Social GAN trained on various dataset generation methods. Training on large amount of sampled data in ``Synth-Large'' achieves the best performance, producing lower prediction errors than training on 100\% real data. Figure~\ref{fig:predictions} shows a qualitative comparison for both of these models. This lower error performance holds true for ``Synth-Large-20\%'' as well, where we sampled from only 20\% of the real data to make the synthetic dataset.
We can attribute part of this high performance to the increased amount of sampled data used for training.
``Synth-Large'' outperforms ``Synth-Equal'', where the only difference is the amount of sampled data (15k more frames in ``Synth-Large''). The realistic variations contained in the additional labeled frames allows for learning a better representation for the true distribution of pedestrian trajectories. This performance difference is especially pronounced in the ADE. We observe similar performance when trained on 100\% versus 20\% real data, where using more training data increases the performance. 
%The difference of over 15k frames is especially pronounced in decreasing the average distance error.

When adding the real dataset to the sampled dataset (in ``Real+Synth-Large''), we do not observe strictly increased performance compared to training on ``Synth-Large'' alone. While the ADE have increased slightly when adding the real dataset, the MDE has decreased.
The lack of large performance increases makes intuitive sense, since the synthetic data is sampled from the real data and the  pedestrian statistics from real data is largely contained in these stochastically sampled trajectories.

In the next section, we will show that adding real data promotes the expression of uncertainty in the DNN, which aids in lowering the MDE. We also show that the small decreases in MDE depend more on the variations in the sampled data than the amount of data sampled through an ablation study.

\begin{table*}[h]
\centering
\caption{Prediction performance across all datasets and methods (best in \textbf{bold} and second best \underline{underlined}). The lower the errors, the better the performance. All errors are reported in meters. The $\mu$ refers to the mean error value across all datasets for each of the ADE, MDE, and FDE evaluation metrics. Each row corresponds to results on a test dataset. For example, the first row reports the ADE values when testing on the \textit{ETH} dataset while trained on the other three datasets.}
\begin{tabular}{ |c|c|c|c|c|c|c|c|c|c| } 
 \hline
  & & \multicolumn{2}{c|}{Real} &  \multicolumn{2}{c|}{Real+Synth-Large}  &  \multicolumn{2}{c|}{Synth-Equal}  & \multicolumn{2}{c|}{Synth-Large}\\
 Metric & Dataset &  20\% &  100\% &  20\%  &  100\%&  20\% & 100\%  &  20\% &  100\% \\
 \hline
        & ETH     &0.95  &0.82   &0.82  &0.77  &0.96  &0.80  & \underline{0.79}        &\textbf{0.75}      \\
 ADE    & Hotel   &0.83  &0.63   &\underline{0.48}  &0.64  &0.70  &0.57  &\textbf{0.43}        &\textbf{0.43}      \\
        & Zara    &0.96  &0.39   &0.36  &0.39  &0.68  &0.37  &\textbf{0.30}        &\underline{0.35}      \\
        & Univ    &0.72  &0.55   &\textbf{0.37}  &\textbf{0.37}  &0.46  &\underline{0.38}  &\underline{0.38}        &\underline{0.38}      \\
\hline
 $\mu$    &         &0.86  &0.60  &0.51 &0.54  &0.70  &0.53  &\textbf{0.47}        &\underline{0.48}      \\
\hline
\hline
        & ETH     &0.56  &0.51   &0.45  &\textbf{0.38}  &0.55  &0.42  &0.45        &\underline{0.40}      \\
 MDE    & Hotel   &0.47  &0.17   &0.15  &\textbf{0.12}  &0.17  &0.16  &\underline{0.13}        &\textbf{0.12}      \\
        & Zara    &0.44  &\underline{0.10}   &0.12  &\textbf{0.09}  &0.14  &0.11  &0.13        &0.11      \\
        & Univ    &0.37  &0.21   &0.18  &\underline{0.15}  &\textbf{0.13}  &\underline{0.15}  &0.20        &0.17      \\
\hline
 $\mu$    &       &0.46  &0.25   &0.23  &\textbf{0.19}  &0.25  &0.21  &0.23     &\underline{0.20}      \\
\hline
\hline
        & ETH     &1.79  &1.61   &1.65 &\underline{1.55} &1.76  &1.56 &1.59  &\textbf{1.50}   \\
 FDE    & Hotel   &1.55  &1.29   &0.99 &1.33 &1.25  &1.10 &\textbf{0.84}  &\underline{0.86}  \\
        & Zara    &1.77  &0.81   &0.75 &0.84 &1.27  &0.75 &\textbf{0.62}  &\underline{0.72}      \\
        & Univ    &1.31  &1.06   &\underline{0.78} &\textbf{0.77} &0.91  &\textbf{0.77} &\underline{0.78}  &\underline{0.78}      \\
\hline
 $\mu$    &       &1.60  &1.19   &\underline{1.04} &1.12  &1.30 &1.05  &\textbf{0.96}  &\textbf{0.96}     \\
\hline
\end{tabular}
\label{table:main_results}
\end{table*}

\subsection{Ablation Study}
In our ablation study, we removed the dataset fitting terms $\sigma_s$ and $\sigma_p$ from the sampler to see their effect on the performance. Upon removing these terms, we sampled from their respective densities without variance, which is equivalent to setting $\sigma_s$ or $\sigma_p$ to zero.

We sampled large datasets from 100\% of the real data following the same procedure as for generating Synth-Large, once with $\sigma_s = 0$ and again with both $\sigma_s = 0, \sigma_p = 0$. We compared these ``reduced models'' to the Synth-Large results with full pedestrian statistics. These reduced models both attain an average ADE of 0.47 m and FDE of 0.95 m, compared to 0.48 m and 0.96 m of the full model. On the other hand, both have higher MDE. Removing $\sigma_s$ from the sampler increases MDE from 0.20 m to 0.22 m; further removing $\sigma_p$ increases this to 0.23 m.

\begin{figure}[h]
\includegraphics[width=0.5\textwidth]{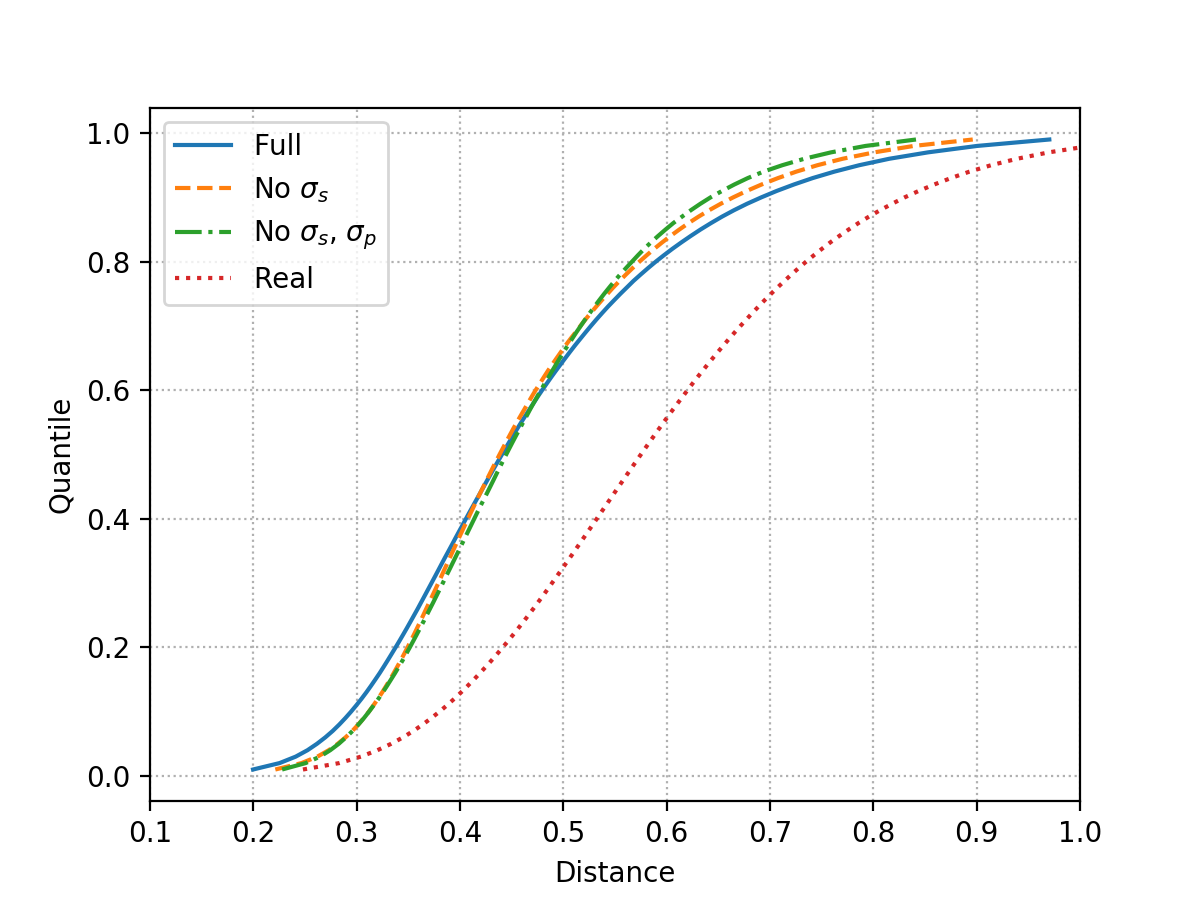}
\centering
\caption{Performance curve for the ablation study using quantile-based metric. The model trained on synthetic trajectory samples has lower distance error than trained on real dataset across all percentiles. Removing the real-pedestrian statistics terms from the sampler reduces the expression of uncertainty.} \label{fig:quantiles}
\end{figure}

We also report an additional quantile-based metric to evaluate their performances. Recall that for each pedestrian at each predicted timestep, Social GAN produced 100 probabilistic predictions. This quantile-based metric is defined by sorting the predictions by their distance to the ground truth pedestrian locations and calculating the average distance for each quantile. Figure~\ref{fig:quantiles} shows the quantile-based distance metric across all training datasets. Ideally, we want the curve to have low distance value across all quantiles. Naturally, the higher the quantile value, the higher the distance (since we sorted the distances in ascending order). The further the curve is shifted towards the top-left, the better the performance. Since we have 100 predictions, the ``quantile'' here is equivalent to ``percentile''.

The lowest point on each curve is equivalent to MDE, since it represents the closest (minimum) distance. The full sampler (with non-zero  $\sigma_s$ and $\sigma_p$ values) performs the best in the 50\% percentile (when quantile less than 0.5 on the plot). The model with no $\sigma_s$, on the hand hand, outperforms that the no $\sigma_s, \sigma_p$ model at the minimum distance as well as in the middle quantiles. All three synthetic-based models outperform the model trained on real data by a great margin across all percentiles.

The more gentle  slope for the real model in Figure~\ref{fig:quantiles}  reflects a greater spread in distances to the ground truth locations as well as uncertainty in the probabilistic predictions.
Expressing this uncertainty is important for safety-critical applications such as autonomous driving, since we would like to avoid hitting any predicted locations a pedestrian may potentially be. Training on the sampled data without real-pedestrian statistics (no $\sigma_s$ and $\sigma_p$) reduces this expression of uncertainty and results in the steeper slopes. Including real-pedestrian statistics and calibrating the sampler to the real data before sampling can help recover this uncertainty, as shown in Figure~\ref{fig:quantiles} by the less steep slopes for the full model, where $\sigma_s$ and $\sigma_p$ were added.

\section{Conclusion} \label{sec:conclusion}

In this work, we presented a novel stochastic sampling method for simulating realistic pedestrian trajectories. We developed a model to extract pedestrian number and walking speed from a small real dataset, and used this information to sample synthetic pedestrian trajectories. We trained a Social GAN on the sampled datasets and evaluated the prediction results on a variety of benchmark datasets of pedestrian trajectories. We show improved prediction performance when trained on large amounts of synthetic data generated by the proposed sampling scheme when compared with trained on real datasets. We also performed an ablation study on the effect of the pedestrian statistics and show that our extracted  pedestrian parameters can represent how pedestrians walk in real dataset and allow the DNN to more accurately model the true distribution of pedestrian trajectories.

Future directions include extending the sampling method to incorporate scene geometry, and training a DNN that utilizes the scene information such as~\cite{manh2018scene} on the synthetic datasets. Sampling from the space of interactions, such as sampling the outcomes of pedestrian yielding, is another direction.

%===============================================================================


%==============================================================================



% \section*{ACKNOWLEDGMENT}

% The preferred spelling of the word �acknowledgment� in America is without an �e� after the �g�. Avoid the stilted expression, �One of us (R. B. G.) thanks . . .�  Instead, try �R. B. G. thanks�. Put sponsor acknowledgments in the unnumbered footnote on the first page.

%==============================================================================
%\clearpage

\bibliographystyle{IEEEtran}
\bibliography{IEEEabrv,root}

\end{document}